\documentclass[letterpaper, 10pt, conference]{ieeeconf}
\IEEEoverridecommandlockouts                              
\let\labelindent\relax
\usepackage{graphicx}
\usepackage{algorithm,algorithmic}
\usepackage{subcaption}
\usepackage{eqlist}
\usepackage{amsfonts}
\usepackage{tabularx}
\usepackage{booktabs}
\usepackage{amssymb}
\usepackage{amsmath}
\usepackage{enumitem}
\usepackage{mathtools}
\usepackage{threeparttable}
\usepackage{lipsum}

\usepackage[usenames,dvipsnames]{xcolor}
\usepackage{comment}
\usepackage{balance}

\usepackage{multirow}

\usepackage{wrapfig}
\usepackage{url}
\newcommand{\figref}[1]{Fig.\ref{#1}}
\newcommand{\tabref}[1]{TABLE \ref{#1}}

\title{\LARGE \bf
Assembly Planning by Recognizing a Graphical Instruction Manual}

\author{Issei Sera$^{1}$, Natsuki Yamanobe$^{2}$, Ixchel G. Ramirez-Alpizar$^{2}$, Zhenting Wang$^{1}$\\ Weiwei Wan$^{1}$, and Kensuke Harada$^{1}$

\thanks{$^{1}$Graduate School of Engineering Science, Osaka University, 1-3 Machikaneyama, Toyonaka, 560-8531, Japan
        {\tt\small \{sera@hlab., ou@hlab., wan@, harada@\}sys.es.osaka-u.ac.jp}}%
\thanks{$^{2}$Automation Research Team, Industrial CPS Research Center, National Institute of Advanced Industrial Science and Technology, 2-4-7 Aomi, Koto-ku, Tokyo, 135-0064, Japan}%
}

\begin{document}

\maketitle
\thispagestyle{empty}
\pagestyle{empty}

\begin{abstract}
This paper proposes a robot assembly planning method by automatically reading the graphical instruction manuals design for humans. Essentially, the method generates an Assembly Task Sequence Graph (ATSG) by recognizing a graphical instruction manual. An ATSG is a graph describing the assembly task procedure by detecting types of parts included in the instruction images, completing the missing information automatically, and correcting the detection errors automatically. To build an ATSG, the proposed method first extracts the information of the parts contained in each image of the graphical instruction manual. Then, by using the extracted part information, it estimates the proper work motions and tools for the assembly task. After that, the method builds an ATSG by considering the relationship between the previous and following images, which makes it possible to estimate the undetected parts caused by occlusion using the information of the entire image series. Finally, by collating the total number of each part with the generated ATSG, the excess or deficiency of parts are investigated, and task procedures are removed or added according to those parts.
In the experiment section, we build an ATSG using the proposed method to a graphical instruction manual for a chair and demonstrate the action sequences found in the ATSG can be performed by a dual-arm robot execution. The results show the proposed method is effective and simplifies robot teaching in automatic assembly.

\end{abstract}

\section{Introduction}
In recent years, the life cycle of products has become shorter. Additionally, manufacturing processes have begun to change; instead of the mass production of one product, high-mix low-volume production is being carried out. A typical form of high-mix low-volume production is cellular manufacturing. However, cellular manufacturing requires considerable labor and is significantly dependent on manpower. To enable robots to perform such manufacturing tasks, without providing detailed instructions of the task, a robot must be able to automatically understand the details of the task and accomplish it based on simple or ambiguous knowledge of the task. Written, oral and illustrated instructions are the most widely used types of task instructions for humans. To adopt these types of task instructions on robots, it is necessary for the robots to understand the meanings of the abstract instructions and convert them into robot-implementable forms.
\begin{figure}[tbp]
\centering
    \includegraphics[width=\linewidth]{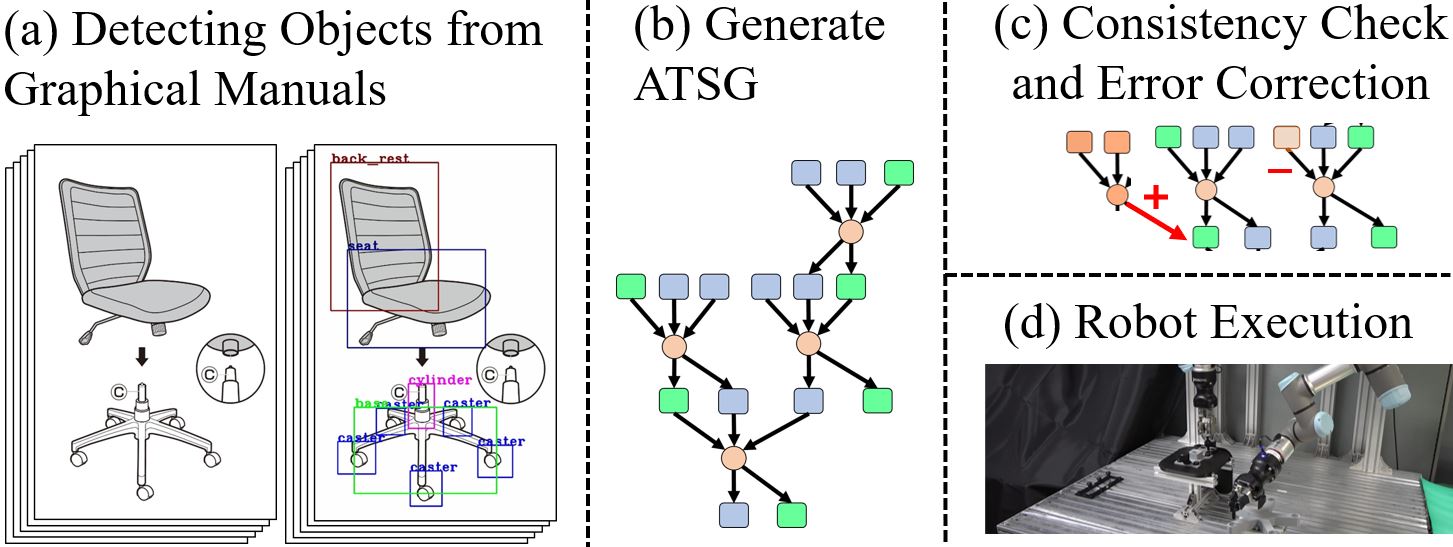}
    \caption{Workflow of the proposed method.}
    \label{fig:overview}
\end{figure}

In conventional research on task planning, instruction extraction is mainly conducted using the linguistic information \cite{interact_picking}\cite{Demo2Text}. Graphical instruction manuals are usually used in the assembly tasks of certain products, such as furniture. However, only a few written task instructions exist in such graphical instruction manuals. Thus in several cases, it is necessary to understand the instructions solely from a series of illustrated instruction images. Information about the parts used for assembly can be obtained from each image included in the illustrated image series, whereas other information about the assembly task itself must be deduced. Moreover, there is no guarantee that the information on the parts is complete.

We proposed a robot assembly planning method based on the illustrated image series of the graphical instruction manual. Fig.\ref{fig:overview} shows the workflow of the method. At the center of the method is an Assembly Task Sequence Graph (ATSG), which is a graph that describes the assembly task procedure that can be executed by a robot. To build an ATSG, the part information is extracted from the instruction images of a graphical instruction manual. Then, the information necessary for planning the assembly task is estimated using the extracted part information. The ATSG includes the relationship between a part and its motion, the task order and the change in the part state owing to that motion, by considering the constraints in the assembly task. First, part information is extracted from each illustrated image included in the illustrated image series of the graphical instruction manual via an object detection system that utilizes deep learning. Next, the information about the assembly task is estimated from the part information, such as motions used to perform the task and the type of tool required to perform the task. This process enables the construction of the unit elements that constitute the ATSG.
Then, the unit elements are integrated by considering the before and after relationships of the image series to generate the ATSG. However, some parts may not be detected in an illustrated image owing to occlusion or occlusion-related conditions. To address this challenge, we estimate the undetected part from the information of the entire image series. Finally, by collating the total number of each part with the ATSG, the excess or deficiency in these parts is evaluated, and task procedures are removed or added.

\section{Related Work}
Robotic assembly is a classical research topic in robotics. Many studies have been conducted to construct assembly sequences based on the geometrical relationships of assembly parts \cite{de1990and}\cite{kwak2011framework}\cite{wilson1994geometric}\cite{Fakhurldeen19}.
More recently, researchers have been focused on combining robotic manipulation planning with assembly task planning. Examples include but are not limited to \cite{roa2015}\cite{dogar2015}\cite{wan2018assembly}\cite{moriyama2019dual}. Compared to these previous assembly studies, this work focuses on recognizing an assembly sequence from a graphical instruction manual and then planning robotic assembly motion. Consequently, we concentrate our review on the generation of robotic task motions via abstracted instructions.

By verbalizing the task contents, a task procedure can be materialized and the reproducibility of the same work by humans and robots can be improved. Previously, Nishimura et al. \cite{Image_Seq} generated instructions by verbalizing a series of task information from a series of cooking images. Erdal et al. \cite{Demo2Text} proposed a framework that could automatically describe task motions from demonstration videos of human tasks. For the study on the generation of robotic task motions via abstracted instructions, Blankenburg et al. \cite{verbal_inst} proposed a framework that generates task sequence graphs from oral instructions in a robot-executable format by considering the constraints of the task order. In this study, the rules of the sentence structure were defined beforehand, and a person verbally instructed each motion or object to be operated according to the defined rules.
Paulius et al. \cite{FOON}\cite{FOON2}\cite{FOON3} modeled cooking tasks based on a human demonstration video. They proposed a knowledge representation called Functional Object-Oriented Network (FOON). The FOON expresses the relationship between motions and objects, and the change in the object state owing to those motions. By using the FOON with the assistance of humans, robots can realize difficult cooking tasks. Beetz et al. \cite{beetz@2017iros} converted cooking recipes into a robot-executable format based on the Action Description. 
The concept of object affordance, which gives an idea about what kind of action can be performed on an object based on its characteristics, has been proposed in \cite{affordance}. Schoeler et al. \cite{tool_affordance} inferred the functional meaning of a tool based on the partial shape of the tool. Understanding the role of objects used in a task is a clue to understanding the meaning of the task. 

Compared to the aforementioned studies, our work is different in that it plans the robotic assembly motion using an ATSG, which is automatically constructed by reading graphical instruction manuals. 
We extract the information on the assembly parts from the instruction images and estimate several feasible information needed for the assembly task such as the assembly motion, assembly tools, and state change of assembly parts. The details of our method will be presented in the following sections.

\section{Assembly Task Sequence Graph (ATSG)}
In this section, we present the background knowledge about ATSG.

\subsection{ATSG Structure}
In the assembly task, multiple separate parts are combined into one while satisfying the constraints derived from the part structure by task motions \cite{And_Or}. The ATSG is a network structure that defines a series of assembly tasks, including the motions of each object that satisfy the constraints. The ATSG comprises assembly units, which are the smallest unit structures in the ATSG. The assembly unit has multiple input object nodes, one motion node and one output object node, and it is connected by directed edges drawn from input object nodes to the motion node and from the motion node to the output object node. One assembly unit is generated for one instruction image in the instructions.

\begin{figure}[!htbp]
\centering
    \includegraphics[width=0.7\linewidth]{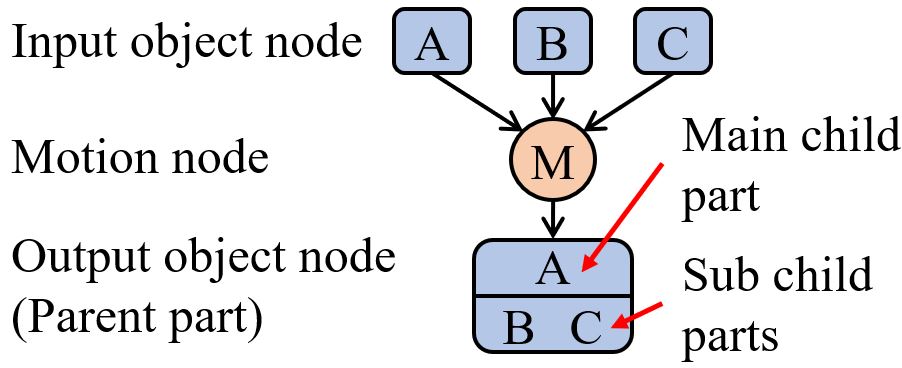}
    \caption{A typical assembly unit of an ATSG.}
    \label{fig:assembly_unit}
\end{figure}

As presented in \figref{fig:assembly_unit}, Part A, B and C are input object nodes, and by passing through the central motion node M, Part A, B and C are combined into one output object node. In other words, each input object node corresponds to a child part, the output object node corresponds to a parent part, and the change in the object state due to the motion is expressed accordingly. The motion is determined from the object affordance of the input object node. The upper part of the output object node is the main child part, whereas the lower part is the subordinate child parts, as illustrated in \figref{fig:assembly_unit}, A is the main child part, and B and C are the subordinate child parts. The name of the output object node is the name of the main child part. By determining the child part information as well as the main-subordinate relationship to the parent part (the output object node), the undetected part can be estimated from the information of the entire image series, even when all parts cannot be detected from the illustrated image owing to the occlusion.

The features of the ATSG are described from three aspects that satisfy the requirements of assembly task sequence planning. The first aspect is the priority for achieving the part-connection relationships. For example, in the assembly task with fasteners, the combination parts are initially installed and then the fasteners are inserted. In addition, because of the branched structure, it is possible to express a task that can be executed in parallel. This makes it possible to determine the work order within a short work time by adopting multiple arms. The second aspect is the usage conditions of the tools. By indicating the tool corresponding to the part being operated in one graph, it is possible to automatically determine the tool that should be used for each task. The third aspect is the task difficulty. By expressing the relationship between the tool and the assembly parts in the entire task with ATSG, for the one-arm task, it is possible to determine the tasks that can be completed continuously without changing the tool. This makes it possible to determine the task order that reduces task difficulty by minimizing the tool change task. In addition, considering the characteristics of robots and humans, the task procedure for inserting fasteners in the instructions is automatically changed, and the number of tasks is reduced.

\subsection{ATSG Generation from Graphical Instruction Manuals}
YOLOv3 \cite{YOLOv3} was used for object detection. Here, the illustrated image model for the graphical instruction manuals of the target product type is learned. Using this model, object detection was conducted on the image series in the instructions. The ATSG is then generated based on the detection results.

As the feature of graphical instruction manuals, there are three problems in using them in robot assembly tasks.
In the ATSG generation, the problems derived from the characteristics of the instructions were solved considering the feasibility of assembly task. By estimating multiple motions, the task order and tools from the relationship of parts, the task of each instruction image is embodied in a particle size that can be realized by robots.

\section{Task Embodiment of Individual Instruction Image}
\label{sec:probrem1}
\begin{figure}[!htbp]
\centering
    \includegraphics[width=\linewidth]{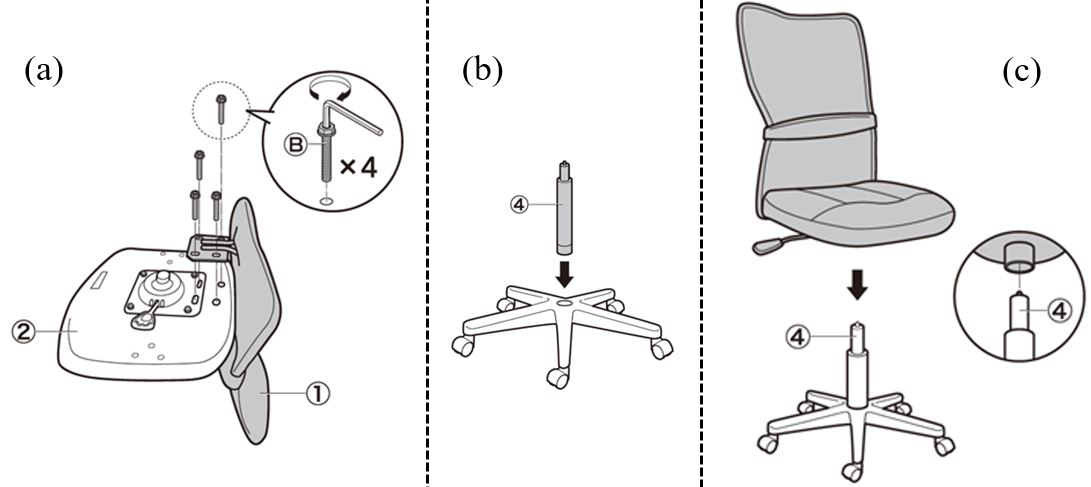}
    \caption{Example of a graphical instruction manual (Instruction image source: \cite{nitori_work_chair}).}
    \label{fig:manual}
\end{figure}

It is difficult to embody the task of each instruction image to particle sizes that can be realized by the robot (Problem 1). Some symbols, such as arrows, are used in the instructions to indicate the assembly procedure.
However, there is no specified operation information in the instructions, and it is often necessary to perform multiple different motions in one instruction image.
For example, in \figref{fig:manual}(a), the following five steps are presented in a single instruction image: Place the back rest the seat in step 1, and use four screws to fix them in steps 2–5. Therefore, it is necessary to estimate the task order and multiple assembly motions from the combination of target parts. 

By estimating the task order, multiple motions, and tools, the task of each instruction image is embodied in a particle size that can be realized by the robot. All these items are estimated from the relationships between objects. There are three major object relationships. The first is the connection relationship between each part. This relationship includes the role of each part in each instruction image and the parts that are combined. For example, in the case of a chair, screws and a seat are parts with fasteners and screw holes, and they are combined. The second relationship is the main-subordinate relationship of the child parts that occurs during the assembly process. The main-subordinate relationship of the child parts in the object node is determined according to the predefined order of each part type based on the role of the part (whether it is a fastener or not) and the part size.

\begin{figure}[!htbp]
\centering
    \includegraphics[width=0.75\linewidth]{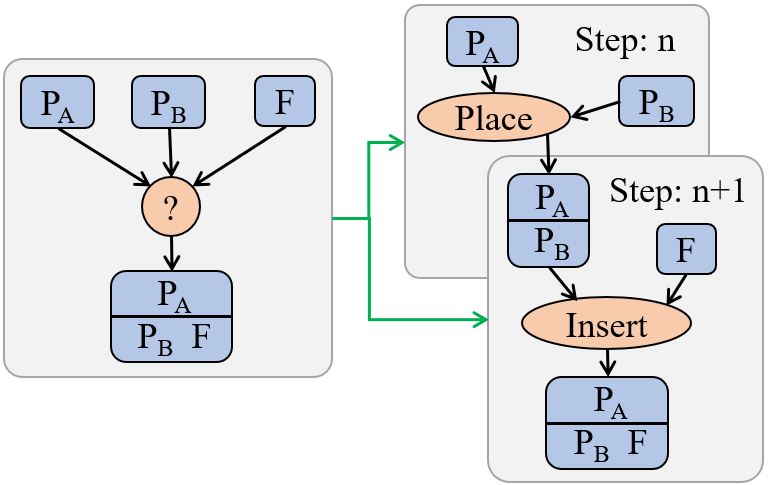}
    \caption{Expansion of an assembly unit.}
    \label{fig:expansion}
\end{figure}

First, the task order is estimated. If there are three or more input object nodes in one assembly unit, by estimating the appropriate task order, the assembly unit expansion is performed to ensure that there are only two input object nodes in one assembly unit. For example, if there are three or more types of parts as input object nodes, including fasteners such as screws and bolts, the assembly unit is expanded to include multiple steps. Tasks on parts other than fasteners are performed in step $n$, whereas tasks on fasteners are assigned to the $n+1$ step. For example, in \figref{fig:expansion}, F is the fastener, ${\rm P_A}$ and ${\rm P_B}$ are parts other than the fastener. In the graphical instruction manual, two steps: (1) placing part ${\rm P_A}$ at the location where it will be combined with ${\rm P_B}$ and (2) inserting fastener F (to fix ${\rm P_A, P_B}$) are often expressed in one instruction image. Under these conditions, the assembly unit was expanded into two steps. 

\begin{figure}[!htbp]
  \begin{center}
    \includegraphics[width=0.87\linewidth]{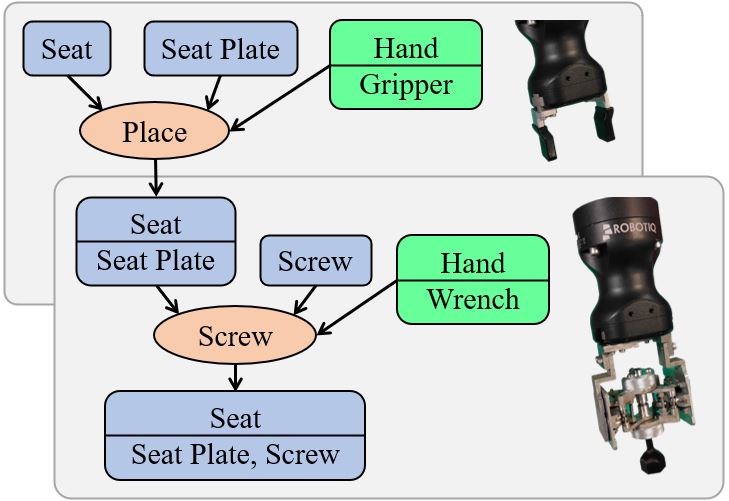}
    \caption{Motion and tool node estimation.}
    \label{fig:motion_hand_node}
  \end{center}
\end{figure}

The next step is the estimation of multiple motions. Based on the concept of object affordance \cite{affordance}, Fukuda et al. \cite{Fukuda_AR} defined the Action relationship. Action relationship generates executable assembly motions from the combination of manipulated target objects. Taking \figref{fig:manual}(a) as an example, it can be estimated that the action of placing the seat plate in a predetermined position of the seat is an action called "place". Furthermore, when inserting a screw into the seat, a motion called "screw" is considered (\figref{fig:motion_hand_node}). 

Finally, the tool is estimated. In the assembly task of robots, it is necessary to adopt a hand or tool with a shape that is suitable for the part to be manipulated. Therefore, when the ATSG is generated, a hand node suitable for the object node is automatically assigned. The hand node is connected by an edge to the same motion node with the corresponding object node. For example, in the assembly of a chair, the hands or tools play two roles: screw tightening and gripping. In the real environment, a dual-arm robot can use one hand holding a screw-tightening tool (wrench) and another hand acting as a gripper, thus the hand node can be illustrated as \figref{fig:motion_hand_node}. By setting the tools according to the part to be manipulated in this way, it is possible to estimate the tools suitable for various parts of each task process when generating the ATSG and planning the task considering actual robot conditions. Therefore, each instruction image information is embodied by expanding the assembly unit while estimating the task information.

\section{Understanding the Relationship between Task Sequence and Extracting Information}
\label{sec:probrem2,3}
It is difficult to determine the relationship between a series of tasks (Problem 2). For example, from \figref{fig:manual}(a), (b), and (c), it is necessary to understand that (c) is the task of combining the parts assembled in (a) and (b).

Furthermore, extracting task information is also difficult (Problem 3). Unlike the demonstration video in the real environment, the instructions, which are presented in images, contain only intermittent task information, and the drawing position and size of the same part also change in different instruction images. Some parts may not be drawn owing to occlusion or incorporation into other parts during the assembly process. For example, in (c) of \figref{fig:manual}, the seat has a seat plate (the part with the lever) and eight screws, However, it is difficult to obtain this information solely from (c).
Therefore, part information needs to be estimated not only from a single instruction image but also from a series of instruction images. Moreover, even if the products are of the same type, the illustration design may differ depending on the manufacturer. Therefore, it is difficult to construct a highly accurate object detection model if a sufficient learning dataset cannot be obtained.

\subsection{Integrating the Graph and Inheriting the Child Part Information}
By connecting the assembly units and inheriting the child part information, it is possible to elucidate the relationship of a series of tasks (Problem 2) and extract the task information from the instructions (Problem 3).
In the instructions, the same type of part may be drawn repeatedly. It is necessary to ascertain whether the part requires additional operations when the same type of part is already assembled in previous instruction images, from the illustrated image series.

\begin{figure}[!htbp]
\centering
    \includegraphics[width=0.7\linewidth]{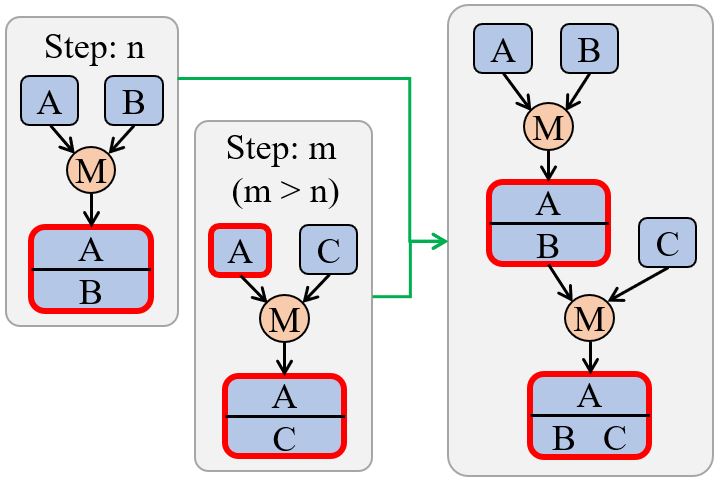}
    \caption{Consolidation and part information inheritance of assembly unit.}
    \label{fig:networking}
\end{figure}

According to the assembly unit expansion rules described in the previous section, all assembly units of each instruction image are expanded such that each of them has only two input object nodes. Subsequently, by referring to the part information shared by the object nodes of all assembly units, it is possible to determine the assembly status of the part detected in each instruction image from the information of previous instruction images, especially whether the part is incorporated in the parent part. This excludes identically created object nodes in all assembly units, which means that the object nodes that have been assembled in the previous instruction image can be excluded. Furthermore, when the output object node matches the input object node in the subsequent steps, the assembly units are connected by integrating them.
For example, as shown in \figref{fig:networking}, the output object node A of the $n$th step and the input object node A of the $m$th step ($m$\textgreater $n$), are the identical parts that exist as duplicates.
Nodes of the same part are searched for and the nodes between the assembly units of the nearest step are integrated.
The subordinate part information of the output object node in the $n$th step is inherited by the final output object node (output object node A at the lower right of \figref{fig:networking}) after concatenation. If the main child part of the output object node in the $n$th step becomes a dependent child part in the $m$th step, all the information of the child parts, including the main child part in the $n$th step, is inherited as the subordinate child part in the $m$th step.

By integrating the image sequence information in this manner, the relationship of a series of tasks is generated as the ATSG while complementing the information that cannot be obtained solely in one instruction image.

\subsection{Error Recovery for Object Detection}
Owing to the challenge of object recognition in instructions, it is not always possible to detect a part with 100\% accuracy. Therefore, even when the object detection result includes false or no detection, the ATSG is complemented by the information of the total number of each part while considering the assembly constraints. By verifying the consistency between the total number of each part and the ATSG final output object node (finished product), it is possible to verify whether the parts included in the finished product have excess or deficiency ((c) of \figref{fig:overview}). If there is an excess or deficiency, the total number of each part and the existing ATSG information are adopted to reconstruct the ATSG while considering the constraints of the assembly task. If there is a missing part in the ATSG, the assembly unit related to the missing part is additionally connected immediately after the assembly step of the same type of part. Next, the ATSG is reconstructed to correct the child parts information and task order. If there are excess parts, the assembly units are removed and the ATSG is reconstructed.

It takes a lot of effort to collect training datasets and build a highly accurate object detection model. By adopting this ATSG complementation method, it is possible to generate an ATSG that can complete the task by using robots even if there is a false detection part. Unless a large number of similar parts is used or the number of parts used in each instruction image is specified, if at least one unassembled part can be detected in each instruction image, it is possible to generate an ATSG that can realize the task.


\section{Experiments and Analysis}
\subsection{Robot Execution Based on Graphical Instruction Manual}
For the assembly task of the robot, the task order, the part to be manipulated, its motion and the required tool were determined according to ATSG.
\subsubsection{Object Detection}
\begin{figure*}[!htbp]
\centering
    \includegraphics[width=0.99\textwidth]{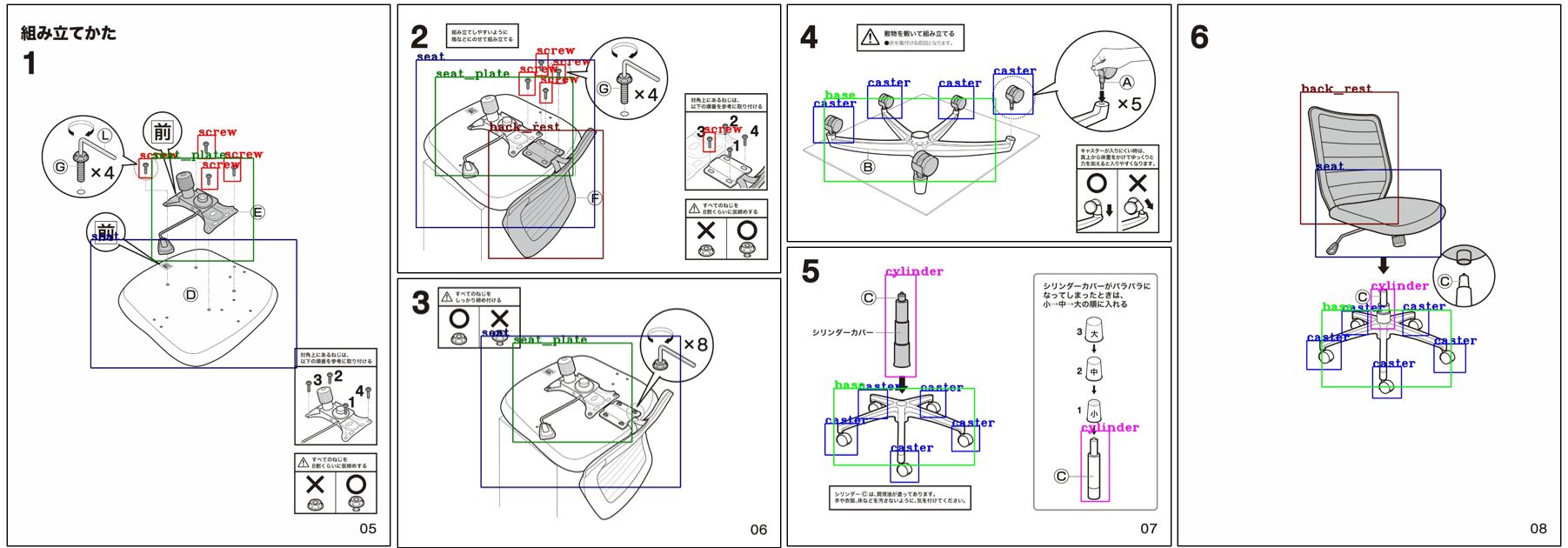}
    \caption{Object detection result of office chair (Instruction image source: \cite{nitori_work_chair}).}
    \label{fig:result_object_detection}
\end{figure*}

\figref{fig:result_object_detection} presents the object detection results obtained from the office chair graphical instruction manual using YOLO v3. The part information for each instruction image extracted by the object detection is presented in \tabref{tab:detected_obj_list}, where the red letters represent falsely detected parts.
In Instruction Image 2 and 5 of \figref{fig:result_object_detection}, the part in the right square area (screw in Instruction Image 2 and cylinder in Instruction Image 5), which is drawn as a supplementary explanation is detected. Because this part is drawn repeatedly in the same instruction image, it is considered a false detection. The parts expected to be detected but not detected are the back rest in Instruction Image 3 and a caster in Instruction Image 4. Most parts with occlusion were not detected because they were incorporated into other parts in the previous instruction images task. For example, only the head of four screws inserted in the seat in Instruction Image 1 is drawn in Instruction Image 2, therefore, it was not detected. Additionally, the seat plate in Instruction Image 6 was not detected.

\begin{table}[!htbp]
    \renewcommand{\arraystretch}{1}
    \centering
    \caption{Detected Parts in \figref{fig:result_object_detection}}
        \begin{threeparttable}
        \begin{tabular}{l|cccccc} \toprule
        Image & 1 & 2 & 3 & 4 & 5 & 6 \\ \midrule
         & Seat & Seat & Seat & Base & Base & Base \\
         & S. P. & S. P. & S. P. & Caster & Caster & Caster \\
         & Screw & Screw &  & Caster & Caster & Caster \\
         & Screw & Screw &  & Caster & Caster & Caster \\
        Part & Screw & Screw &  & Caster & Caster & Caster \\
         & Screw & Screw &  &  & Caster & Caster \\
         &  & \textcolor{red}{Screw} &  &  & Cylinder & Cylinder \\
         &  & B. R. &  &  & \textcolor{red}{Cylinder}  & Seat \\
         &  &  &  &  &  & B. R.\\ \bottomrule
        \end{tabular}
        \begin{tablenotes}
        \item[*] Candidate parts and their ground-truth total numbers in the manual: Base $\times$1; Base Rest $\times$1; Caster $\times$5; Cylinder $\times$1; Screw $\times$8; Seat Plate $\times$1. Abbreviations: S. P. = Seat Plate; B. R. = Back Rest. 
        \end{tablenotes}
        \end{threeparttable}
        \label{tab:detected_obj_list}
\end{table}

\begin{figure*}[!htbp]
\centering
    \includegraphics[width=.99\textwidth]{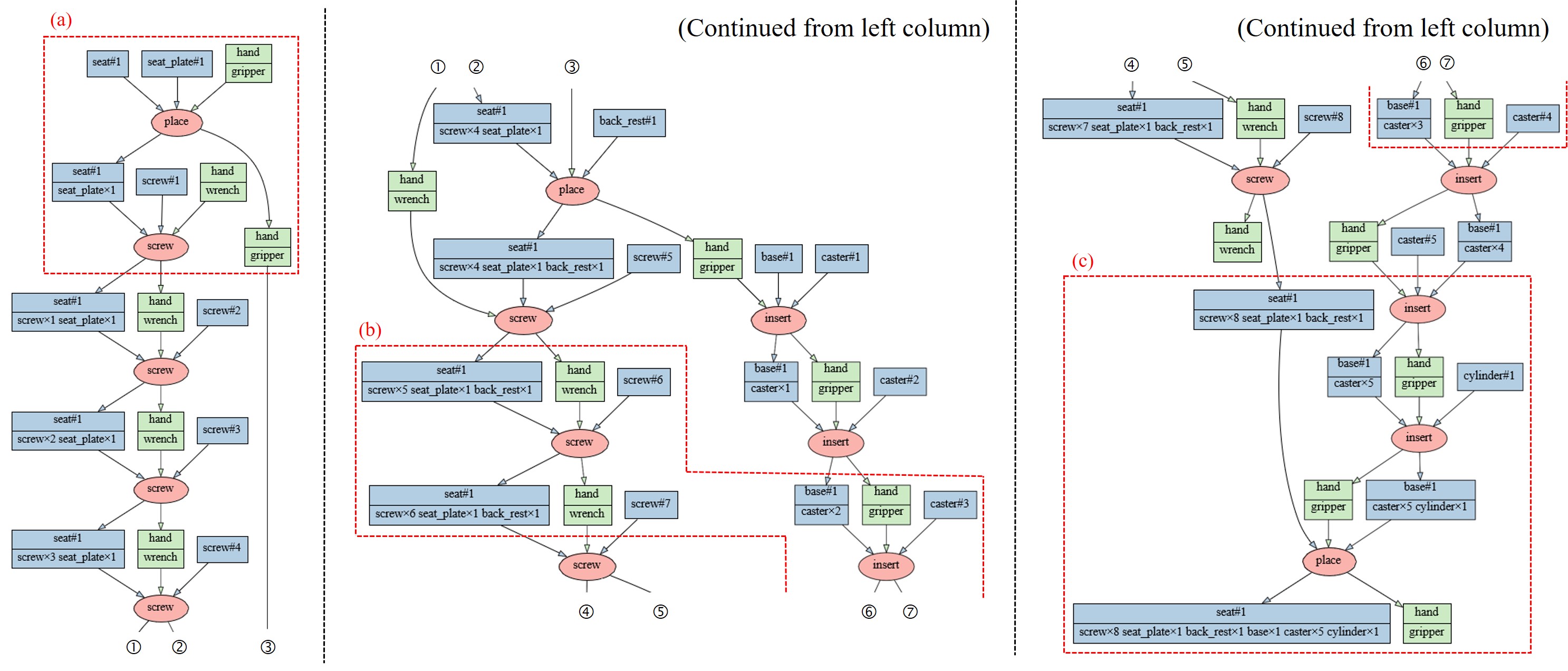}
    \caption{The generated ATSG based on the object detection results (\tabref{tab:detected_obj_list}).}
    \label{fig:atsg_fixed}
\end{figure*}

\begin{figure}[!htbp]
\centering
    \includegraphics[width=\linewidth]{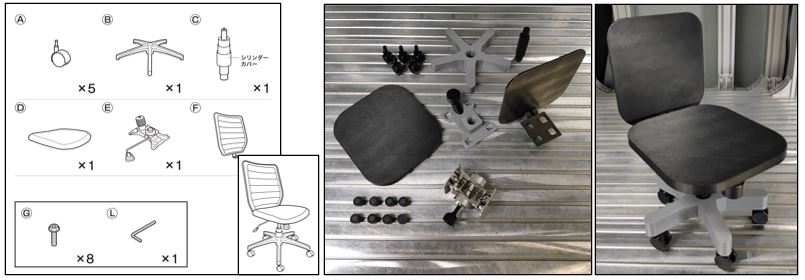}
    \caption{Office chair model produced by 3D printer (right figure). On the left figure are images \cite{nitori_work_chair} of the finished product and each part.}
    \label{fig:chair_model}
\end{figure}

\begin{figure*}[!htbp]
  \centering
    \includegraphics[width=0.99\linewidth]{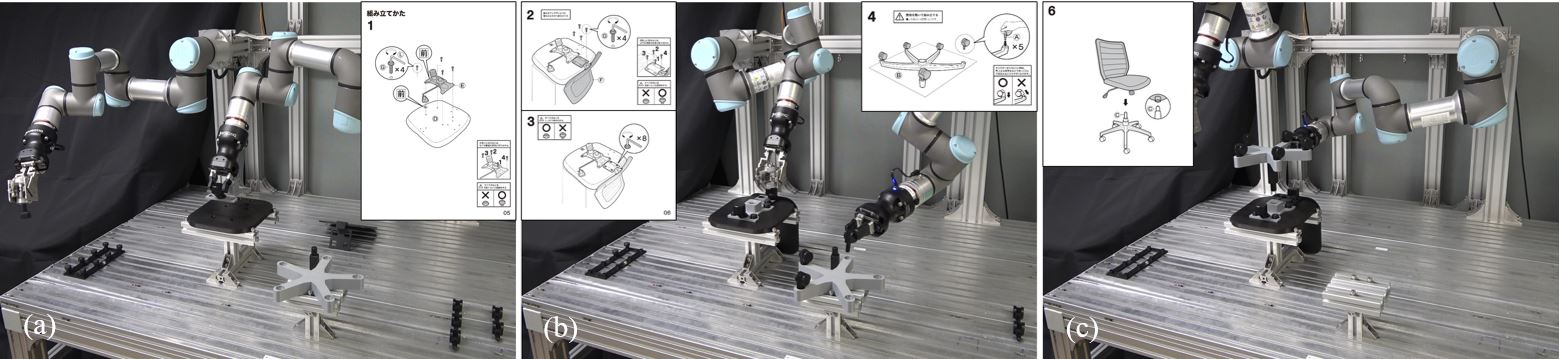}
    \caption{Robot execution. The (a-c) sub-figures correspond to assembly denoted by the red boxes in \figref{fig:atsg_fixed}.}
    \label{fig:rbt_exp}
\end{figure*}
%
\subsubsection{ATSG Generation}
\figref{fig:atsg_fixed} presents the generated ATSG based on the object detection results (\tabref{tab:detected_obj_list}). In this figure, the blue, green and red nodes denote the object, hand and motion nodes, respectively. The ATSG branches in the central part and branches are recombined. This means that the parts (seat part and base part) assembled in Instruction Image 3 and 5 of \figref{fig:result_object_detection} are combined in Instruction Image 6. From the branching structure, the processes corresponding to Instruction Image 1 to 3 and the processes corresponding to Instruction Image 4 to 5 can be executed in parallel.

\tabref{tab:exp_node_comp} presents the node configuration for each instruction image and the entire ATSG. This table can be used to evaluate whether it is embodied in a feasible particle size that can be realized by robots (Problem 1 in Section \ref{sec:probrem1}). From the number of motion nodes in Instruction Images 1, 2 and 4, it can be observed that there are multiple motions in one instruction image. In other words, the task of one instruction image is embodied by multiple motions. In Instruction Image 3, screw-tightening instructions are obtained for the parts in Instruction Image 2. Therefore, because the new assembly part is not illustrated in Instruction Image 3, the node corresponding to Instruction Image 3 is not generated. This indicates that when the same part is detected, the necessity of an additional task is determined by integrating the information of multiple instruction images. The total number of object and hand nodes in all instruction images does not match the number of object and hand nodes in the entire ATSG. This is because the same object nodes or hand nodes in each instruction image are connected by the connection of the assembly units, and one node may be an element in multiple instruction images, which means that the task information of multiple instruction images is integrated.

\begin{table}[!htbp]
    \centering
    \caption{Node Configurations for Each Instruction Image}
    \begin{tabular}[t]{lccc}\toprule
         & Object Node & Motion Node & Hand Node \\ \midrule
        Image 1 & 11 & 5 & 7 \\
        Image 2 & 11 & 5 & 6 \\
        Image 3 & 0 & 0 & 0 \\
        Image 4 & 11 & 5 & 6 \\
        Image 5 & 3 & 1 & 2 \\
        Image 6 & 3 & 1 & 2 \\ \midrule
        Entire ATSG & 25 & 17 & 19 \\ \bottomrule
    \end{tabular}
    \label{tab:exp_node_comp}
\end{table}

\begin{table}[!htbp]
    \centering
    \caption{Number of Complementary Parts for Each Instruction Image}
    \begin{tabular}[t]{@{\extracolsep{3pt}}lcccc} \toprule
         & \multicolumn{2}{c}{Part Information} & \multicolumn{2}{c}{Total Number of Each Part} \\ \cmidrule{2-3} \cmidrule{4-5}
                & Add & Remove & Add & Remove \\ \midrule
        Image 1 & 0   & 0      & 0   & 0 \\
        Image 2 & 4   & 1      & 0   & 1 \\
        Image 3 & 9  & 2      & 0   & 0 \\
        Image 4 & 0   & 0      & 0   & 0 \\
        Image 5 & 0   & 4      & 0   & 1 \\
        Image 6 & 9  & 7      & 0   & 0 \\ \midrule
        All Images& 22  & 14      & 0   & 2 \\ \bottomrule
    \end{tabular}
    \label{tab:exp_comp_obj}
\end{table}

\tabref{tab:exp_comp_obj} presents the number of complemented assembly parts for each instruction image. From this table, we evaluate the feasibility of addressing the challenge in elucidating the relationship of a series of tasks and extracting the information from the instructions (Problem 2 and 3 in Section \ref{sec:probrem2,3}). It can be observed that the child part information of the object nodes is inherited, and the parts that cannot be detected because of occlusion are complemented. For example, the value of Part Information Add Instruction Image 2 indicates that four incorporated screws in Instruction Image 1 are added as part information. In addition, the assembled parts are excluded from the drawing targets as input object nodes, even if they are detected. For example, the value of Part Information Removed of Instruction Image 6 indicates that all parts other than the seat and base assembled in Instruction Image 1-5 are not illustrated as input object nodes, and no additional task for the assembled parts in the previous instruction images is required. At the point of the ATSG reconstruction, the screws in Instruction Image 2 and cylinders in Instruction Image 5, which are false detection parts, are removed based on the total number of each part (Removed of Image 2 and 5 in \tabref{tab:exp_comp_obj}). From these, we confirmed that it is possible to understand the relationship of a series of operations and the proposed method can be used under conditions where extracting the information is challenging.

The ATSG was also generated for the color box and steel rack based on the manually created object detection results. In the experiment, only minimal information was provided as the object detection result, and it was assumed that all the assembled parts in previous instruction images are not detected. By complementing the task information from the image series, we generated an ATSG that could be assembled successfully.

\subsubsection{Robot Execution}
According to the ATSG in \figref{fig:atsg_fixed}, the product model was assembled using a robot. The UR3e robot from Univeral Robot Ltd. was used in the experiment. The screw-tightening tool in \cite{hu2020mechanical} was used.
RRT-connect \cite{RTT-connect} was adopted to generate the route of the assembly operation. Because it is difficult to operate a real office chair with UR3e, we created a pseudo model using a 3D printer, as illustrated in figure \figref{fig:chair_model}. The office chair was successfully assembled by performing an assembly task based on the generated ATSG.

\figref{fig:rbt_exp} presents the robot experiment. The (a-c) sub-figures correspond to assembly denoted by the red boxes in \figref{fig:atsg_fixed}. The execution results are consistent with the ATSG and the instruction image. In (a), the instruction image information is expanded to multiple motions, and the appropriate motion "place" is determined from the combination of the input object nodes (seat and seat plate nodes at the top). In addition, the output object node (center left seat node) holds the placed parts as subordinate child part information (seat plate), and the screw is inserted into the seat in the next step. Furthermore, the seat plate and screw are assigned a gripper and wrench hand node, respectively. The ATSG determines the tool that the robot should use. In (b), the sixth screw is screwed in the seat, and simultaneously, the third caster is inserted into the base. These tasks can be performed independently and using different hands benefiting from the branching structure of the ATSG. Therefore, when multiple hands can be used, work time can be shortened by performing the task in parallel based on ATSG. In (c), the seat and base parts, which can be assembled independently, are finally combined into one. The final output object node holds all the parts as the child part information without excess or deficiency. Therefore, even if an error occurs in the object detection result, a task plan that can be realized by robots can be generated.

\subsection{Error Recovery for Object Detection}
To address the problem of incomplete object detection, we intentionally created a table on the object detection results obtained from the illustration images that are identical to the source images of \figref{fig:result_object_detection}, including the undetected parts (\tabref{tab:wrong_result_object_detection}). Only the parts written in black were used for the ATSG generation. Although the parts written in red are presented in the illustrated image, they are assumed to be undetected and are not used for ATSG generation. The ATSG generated based on \tabref{tab:wrong_result_object_detection} is identical to the ATSG in \figref{fig:atsg_fixed}. From these results, we confirmed that a robotic assembly task can be realized by incomplete object detection.

\begin{table}[!htbp]
\centering
    \caption{Manually Created Object Detection Result including Undetected Parts.}
    \begin{threeparttable}
        \begin{tabular}[t]{l|cccccc}\toprule
        Image & 1 & 2 & 3 & 4 & 5 & 6 \\ \midrule
               & Seat       & Seat                        & \textcolor{red}{Seat}       & Base                    & Base                    & Base \\
               & S. P. & \textcolor{red}{S. P.} & \textcolor{red}{S. P.} & Caster                  & \textcolor{red}{Caster} & \textcolor{red}{Caster} \\
               & Screw      & Screw                       & \textcolor{red}{Screw}      & \textcolor{red}{Caster} & \textcolor{red}{Caster} & \textcolor{red}{Caster} \\
               & Screw      & Screw                       & \textcolor{red}{Screw}      & \textcolor{red}{Caster} & \textcolor{red}{Caster} & \textcolor{red}{Caster} \\
               & Screw      & Screw                       & \textcolor{red}{Screw}      & \textcolor{red}{Caster} & \textcolor{red}{Caster} & \textcolor{red}{Caster} \\
        part & Screw      & Screw                       & \textcolor{red}{Screw}      & \textcolor{red}{Caster} & \textcolor{red}{Caster} & \textcolor{red}{Caster} \\
               &            & \textcolor{red}{Screw}      & \textcolor{red}{Screw}      &                         & Cylinder                & \textcolor{red}{Cylinder} \\
               &            & \textcolor{red}{Screw}      & \textcolor{red}{Screw}      &                         &                         & Seat \\
               &            & \textcolor{red}{Screw}      & \textcolor{red}{Screw}      &                         &                         & \textcolor{red}{S. P.} \\
               &            & \textcolor{red}{Screw}      & \textcolor{red}{Screw}      &                         &                         & \textcolor{red}{B. R.} \\
               &            & B. R.                   & \textcolor{red}{B. R.}  & & &  \\ \bottomrule
        \end{tabular}
    \begin{tablenotes}
    \item[*] Candidate parts and abbreviations are the same as \tabref{tab:detected_obj_list}.
    \end{tablenotes}
    \end{threeparttable}
        \label{tab:wrong_result_object_detection}
\end{table}



\section{Conclusions and Future Work}
In this study, we proposed a method to generate an ATSG by extracting part information from an illustrated image series of a graphical instruction manual. The generated ATSG can deduce the task motions and their orders, assembly state, and tools. Additionally, the generated ATSG can reconstruct and express instructions for humans into a form that is executable by robots.

In the future work, we will automatically generate robot motions from the ATSG, and consider the incorporation of the environmental information into the ATSG. Since task and motion planning (TAMP) depends on the work environment, it is necessary to integrate the ATSG with environmental information such as the location of the parts and tools in motion planning. Furthermore, directly obtaining task motions and object relationship information from a graphical instruction manual, and applying the ATSG to a variety of tasks that use instruction manuals, such as industrial products in other domains and cooking, are expected.


\balance
\bibliography{root}
\bibliographystyle{unsrt}

\end{document}